\documentclass{article}

\usepackage{soul}

\usepackage[final,nonatbib]{neurips_2023_ml4ps}

\usepackage[utf8]{inputenc} 
\usepackage[T1]{fontenc}    
\usepackage{hyperref}       
\usepackage{url}            
\usepackage{booktabs}       
\usepackage{amsfonts}       
\usepackage{nicefrac}       
\usepackage{microtype}      
\usepackage{xcolor}         
\usepackage{graphicx}
\usepackage{amsmath,amssymb}
\usepackage{comment}
\usepackage{mfirstuc}

\newcommand{\addComment}[2]{
  \expandafter\newcommand\csname #1\endcsname[1]{{\bf \color{#2} \capitalisewords{#1}:\,##1}}
  \expandafter\newcommand\csname #1cor\endcsname[2]{{\color{#2} \capitalisewords{#1}:\,\st{##1}{\bf ##2}}}
  \expandafter\newcommand\csname #1color\endcsname{#2}
}
\addComment{cris}{blue} 
\addComment{james}{red}

\title{Uncertainty Quantification with Bayesian Higher Order ReLU KANs}

\author{%
  James Giroux, Cristiano Fanelli\\
  Department of Data Science\\
  William \& Mary\\
  Williamsburg, VA 23185 \\ 
  \texttt{\color{blue}\{jgiroux, cfanelli\}@wm.edu} \\
}

\begin{document}

\maketitle

\begin{abstract}
We introduce the first method of uncertainty quantification in the domain of Kolmogorov-Arnold Networks, specifically focusing on (Higher Order) ReLUKANs to enhance computational efficiency given the computational demands of Bayesian methods. The method we propose is general in nature, providing access to both epistemic and aleatoric uncertainties. It is also capable of generalization to other various basis functions. We validate our method through a series of closure tests, including simple one-dimensional functions and application to the domain of (Stochastic) Partial Differential Equations. Referring to the latter, we demonstrate the method's ability to correctly identify functional dependencies introduced through the inclusion of a stochastic term. The code supporting this work can be found at \href{https://github.com/wmdataphys/Bayesian-HR-KAN}{https://github.com/wmdataphys/Bayesian-HR-KAN}.
\end{abstract}

\section{Introduction}

The introduction of Kolmogorov-Arnold Networks (KANs) \cite{liu2024kan} as substitutes for more traditional Multi-Layer Perceptron (MLP) has been seen as a form of revolution in the deep learning community. MLP's, backed by the universal approximation theorem \cite{hornik1989multilayer}, deploy functional learning directly through a set of weight matrices. As such, the `activation' function applied at each layer must be chosen apriori. In contrast, KAN's, backed by the Kolmogorov-Arnold representation theorem \cite{Kolmogorov_1961}, learn both the weights and the functional form of the activation function at training time. Specifically, KAN's rely on the fact that any continuous multivariate function $f$, operating on a bounded domain can be written as a finite sum of continuous univariate functions, Eq. \ref{eq:f_comp}. \cite{liu2024kan}

\begin{equation}\label{eq:f_comp}
    f(\bold{x}) = \sum_{q=1}^{2n + 1} \Phi_q \left(\sum_{p=1}^{n} \phi_{q,p}(x_p) \right), \; \phi_{q,p}: [0,1] \rightarrow \mathbb{R}, \; \Phi_{q}: \mathbb{R} \rightarrow \mathbb{R}
\end{equation}

The above representation is limiting due to limited non-linearities (2) and small number of terms ($2n + 1$). Liu et al. \cite{liu2024kan} remove this limitation through the realization of the relation to MLPs, in which a single KAN layer, Eq. \ref{eq:kan_layer}, with n-dimensional inputs, can be defined as a matrix of one-dimensional functions, capable of being chained together to form a network of any depth.


\begin{equation}\label{eq:kan_layer}
    \bold{x}_l =
\begin{pmatrix}
\phi_{l,1,1}(\cdot) & \phi_{l,1,2}(\cdot) & \cdots & \phi_{l,1,n_l}(\cdot) \\
\phi_{l,2,1}(\cdot) & \phi_{l,2,2}(\cdot) & \cdots & \phi_{l,2,n_l}(\cdot) \\
\vdots & \vdots &  & \vdots \\
\phi_{l,n_{l+1},1}(\cdot) & \phi_{l,n_{l+1},2}(\cdot) & \cdots & \phi_{l,n_{l+1},n_l}(\cdot)
\end{pmatrix} \bold{x}_{l-1}
\end{equation}

In Liu et al. \cite{liu2024kan}, the activation functions $\phi(x)$, are chosen to be linear combinations of a basis function and a B-Splines. 

\section{Related Works}

Since the introduction of KAN, many improvements have been made both from computational standpoints, but also choices of expressive basis functions. \cite{eff_KAN,li2024kolmogorovarnoldnetworksradialbasis,Athanasios2024,ta2024bsrbfkancombinationbsplinesradial,bozorgasl2024wavkanwaveletkolmogorovarnoldnetworks,ss2024chebyshevpolynomialbasedkolmogorovarnoldnetworks,yang2024activationspaceselectablekolmogorovarnold} In \cite{qiu2024relu}, the authors introduce \textit{ReLU-KAN}, in which the choice of basis functions is replaced with those more suitable for GPU parallelization. Specifically, the functions require only matrix addition, dot products and ReLU activations. They replace the B-Spline functions in \cite{liu2024kan} with combinations of ReLU functions and a normaliation term, Eq. \ref{eq:relu_kan}, where $s_i$ and $e_i$ are learnable parameters defining the start and end of the domains.




\begin{equation}\label{eq:relu_kan}
    R_i(x) = [\text{ReLU}(e_i - x) \times \text{ReLU}(x - s_i)]^2 \times 16 / (e_i - s_i)^4
\end{equation}

$\phi(x)$ can be represented through Eq. \ref{eq:relu_kan_phi}, where $G$ and $k$ are the number of grids and the span parameter, respectively. The Grid parameter $G$ controls the number of functions contributing to individual activations, and the span parameter $k$ parameterizes the non-zero region of individual basis functions. 

\begin{equation}\label{eq:relu_kan_phi}
    \phi(x) = \sum_{i=1}^{G+k}w_iR_i(x)
\end{equation}

As described in Qiu et al. \cite{qiu2024relu}, the input/output scheme of a input vector $\bold{x} = \{x_1,x_2, \hdots , x_n \} $ and output vector $\bold{y} = \{y_1,y_2, \hdots , y_m \} $ can be decomposed into singular matrix operation:


\begin{equation}\label{eq:conv}
    \bold{y} = \bold{W} \otimes \begin{pmatrix}
R_1(x_1) & R_2(x_1) & \cdots & R_{G+k}(x_1) \\
R_1(x_2) & R_2(x_2) & \cdots & R_{G+k}(x_2) \\
\vdots & \vdots & \ddots & \vdots \\
R_1(x_n) & R_2(x_n) & \cdots & R_{G+k}(x_n)
\end{pmatrix}
\end{equation}

Where $\otimes$ denotes the convolution operation. The result is a more computationally efficient and expressive set of a basis functions. However, as pointed out in So et al. \cite{SO_2024}, the smoothness of higher order derivatives of the perviously defined basis functions is a highly limiting factor, specifically for Partial Differential Equations (PDEs) that posses higher order derivatives. As a result, they provide a more general framework, in which higher order basis functions (opposed to the square of ReLUs) are represented in Eq. \ref{eq:higher_order_rkan}.

\begin{equation}\label{eq:higher_order_rkan}
    R_{i,m}(x) = [\text{ReLU}(e_i - x) \times \text{ReLU}(x - s_i)]^m \times (2 / (e_i - s_i))^{2m}
\end{equation}


\section{The Bayesian Approach}

Inspired by Bayesian Neural Networks (BNNs), we aim to define posterior distributions, $q(\bold{W}|D)$, over the parameter space of KAN, allowing prediction through a posterior distribution $q(\bold{y}|\bold{x},D)$, integrated over the space of the weights. In reality, such a posterior is intractable and one must utilize Bayesian inference techniques during optimization. We define posteriors over the basis functions $R_{i,m}(\bold{x})$, at both second (default ReLU-KAN) and higher order ($m > 2$) such that we can efficiently account for the uncertainty in our network. Specifically, we define posteriors over the parameters $\bold{e}_i,\bold{s}_i$ through variational inference by introducing two new learnable parameters $\sigma(\bold{e}_i), \sigma(\bold{s}_i)$:

\[
\begin{aligned}
\hat{\bold{s}_i} &= \mu(\bold{s}_i) + \sigma(\bold{s}_i) \odot \epsilon \\
\hat{\bold{e}_i} &= \mu(\bold{e}_i) + \sigma(\bold{e}_i) \odot \epsilon
\end{aligned}
\]

Where $\epsilon$ is represented by a fully factorized Gaussian. In essence, we are defining a posterior over the start and end of the unary bell shaped basis functions. Furthermore, we wish to parameterize the weights of the convolution ($\bold{W}$ of Eq. \ref{eq:conv}) making the posterior over the two operations joint. To do so, we utilize the convolutional layers defined in \cite{pmlr-v70-louizos17a}. In contrast to the fully-factorized assumption on $q(\bold{W})$ made prior, which can limit inherent network complexity, the authors instead represent the posterior distribution through the product of a fully-factorized Gaussian and a mixing density, Eq. \ref{eq:mixing}.

\begin{equation}\label{eq:mixing}
    q(\bold{W}) = \int q(\bold{W} | \bold{z}) q(\bold{z}) d\bold{z}
\end{equation}

Where $q(\bold{z})$ is a set of random variables acting only multiplicatively on the means of $q(\bold{W} | \bold{z})$ to reduce computational overhead. The resulting posterior distribution over the weights becomes more flexible, capable of capturing complex multi-modal dependencies. However, this leads to intractability, necessitating the construction of an approximate lower bound for the entropy. This is achieved using an auxiliary distribution, \( r(\bold{z}|\bold{W}) \), which is equivalent to performing variational inference in an augmented probability space \cite{pmlr-v70-louizos17a}. Importantly, the approximate posterior remains valid due to the fact that the auxiliary distribution can be marginalized out. As \cite{pmlr-v70-louizos17a} state, the tightness of the bound on \( q(\bold{W}) \) — and thus its quality — depends on how well \( r(\bold{z}|\bold{W}) \) approximates the posterior \( q(\bold{z}|\bold{W}) \). To achieve this, inverse normalizing flows are employed. During training, the approximate posterior is constrained by Eq. \ref{eq:KL}, serving as a regularization term alongside the standard loss function.

\begin{align}\label{eq:KL}
    \mathcal{L}_{KL.} &= -KL(q(\bold{W})\|p(\bold{W})) \nonumber \\
    &= \mathbb{E}_{q(\bold{W},\bold{z}_T)}[-KL(q(\bold{W}|\bold{z}_{T_f}) \| p(\bold{W})) \nonumber \\
    & + \log r(\bold{z}_{T_f} | \bold{W}) - \log q(\bold{z}_{T_f})]
\end{align}

The above formulation allows estimation of the epistemic uncertainty (\textit{i.e.}, model uncertainty) at inference through sampling of the posterior. However, in the space of physical problems in which inputs are injected from physical processes, aleatoric uncertainty (\textit{i.e.}, stochastic uncertainty inherent to the data) must be accounted for. In \cite{kendall2017uncertainties}, the authors propose directly learning the aleatoric uncertainty at training time through a Gaussian Log-likelihood, in which the output of the network $\bold{y} = \{y_1,y_2, ..., y_n\}$ is augmented with the associated uncertainty $\sigma(\bold{y}) = \{\sigma(y_1), \sigma(y_2), ..., \sigma(y_n)\}$, learned through Eq. \ref{eq:gauss_lkd}, where $\phi_{\bold{W}}$ denotes parameterization through a model. In our case, the output of the network is one-dimensional, resulting in easy computation and no consideration of covariance.

\begin{equation}\label{eq:gauss_lkd}
    \mathcal{L}(\bold{u} | \bold{x},\bold{\hat{u}}_{\phi_\bold{W}},\boldsymbol{\hat{\sigma}}_{\phi_\bold{W}}) = \frac{1}{N}\sum_j^{N} \frac{1}{2}( e^{-\bold{r_j}} \|\bold{u}_j - \bold{\hat{u}}_j\|^2 + \bold{r}_{j}), \; \bold{r}_j = \log \boldsymbol{\sigma}_j^2
\end{equation}

\begin{figure}
    \centering
    \includegraphics[width=0.75\textwidth, trim={8.0cm, 0.0cm, 1.0cm, 0.0cm}, clip]{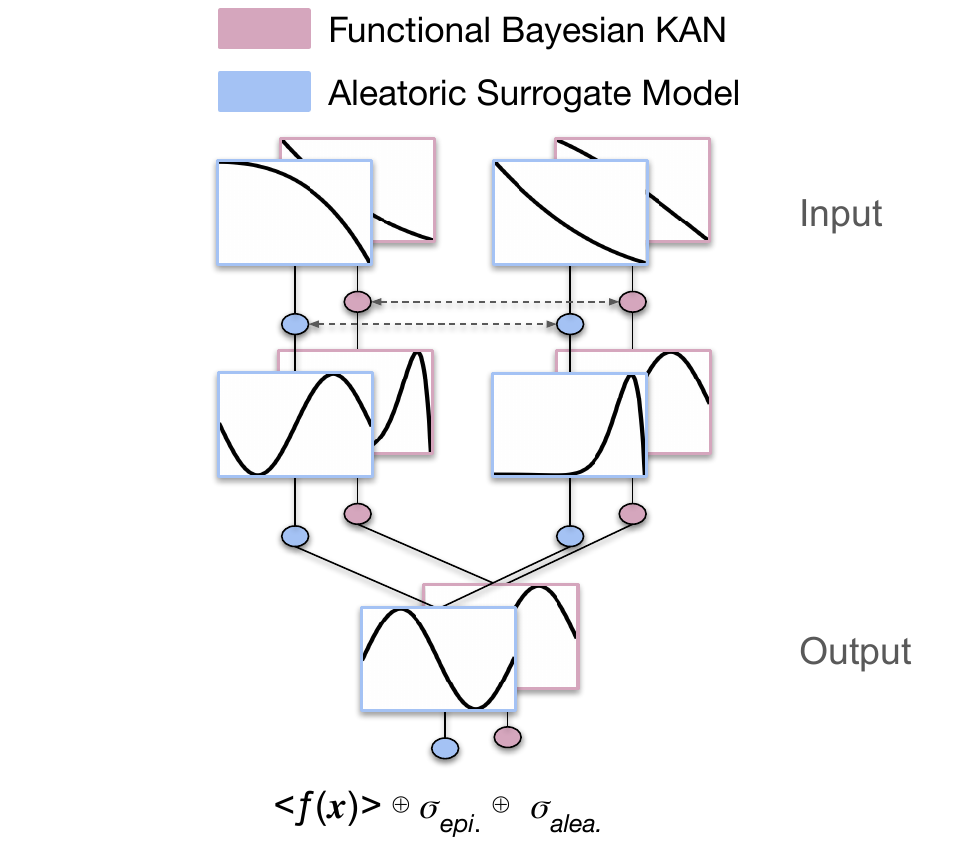}
    \caption{\textbf{Bayesian (Higher Order) ReLU KAN:} The method consists of two Bayesian (Higher Order) ReLU KANs, in which we define a surrogate model tasked with learning the aleatoric component in the likelihoods. This model is capable of inheriting the structure of the functional KAN, or with its own unique set of parameters.}
    \label{fig:bkan_flowchart}
\end{figure}

In the case of BNNs, this is often represented through a bicephalous network, in which the output is formulated into two disjoint heads. \cite{fanelli2024eluquant,giroux2024unmasking} We instead represent the aleatoric uncertainty through a surrogate model, taking the form of another Bayesian KAN, depicted in Fig. \ref{fig:bkan_flowchart}.

This surrogate KAN inherits its structure from the functional KAN by default, although we encode the ability to define separate structures. The loss is then given by the summed contributions of Eq. \ref{eq:KL} and \ref{eq:gauss_lkd}. 
 We also provide code for fitting through a Student-t distribution, Eq. \ref{eq:student_t}, in which we introduce another learnable parameter $\nu$. Such a likelihood may be more robust to outliers, although this is generally data-dependent and inherently tied to the complexity of the problem. In simple cases, a Gaussian assumption is often sufficient even under the presence of strong outliers.


\begin{equation}\label{eq:student_t}
\begin{aligned}
    \mathcal{L}(\bold{u} | \bold{x},\bold{\hat{u}}_{\phi_\bold{W}},\boldsymbol{\hat{\sigma}}_{\phi_\bold{W}},\hat{\nu}_{\phi_\bold{W}}) = & \frac{1}{N}\sum_j^N -\log{\Gamma(\frac{\nu + 1}{2})} + \log{\Gamma(\frac{\nu}{2})} + \log{\sqrt{\nu \pi \boldsymbol{\sigma}_j^2}} \\
     & +\frac{\nu + 1}{2}\log{\left( 1 + \frac{||\bold{u}_j - \bold{\hat{u}}_j||^2}{\nu\boldsymbol{\sigma}_j^2}\right)}
\end{aligned}
\end{equation}

\section{Experiments}

In the following, we demonstrate the the ability of the Bayesian model on function fitting, along with the ability to extract meaningful representations of both the epistemic, and aleatoric uncertainties. 

\subsection{One Dimensional Fits}

We follow the experiments in \cite{qiu2024relu}, selecting set of one-dimensional functions given below:

\begin{equation}\label{eq:list_of_funcs}
    \begin{aligned}
        f_1(x) & = \sin(\pi x) \\
        f_2(x) & = \sin(5\pi x) + x \\
        f_3(x) & = e^{x} 
    \end{aligned}
\end{equation}

\begin{figure}[h]
    \centering
    \includegraphics[width=0.3\textwidth]{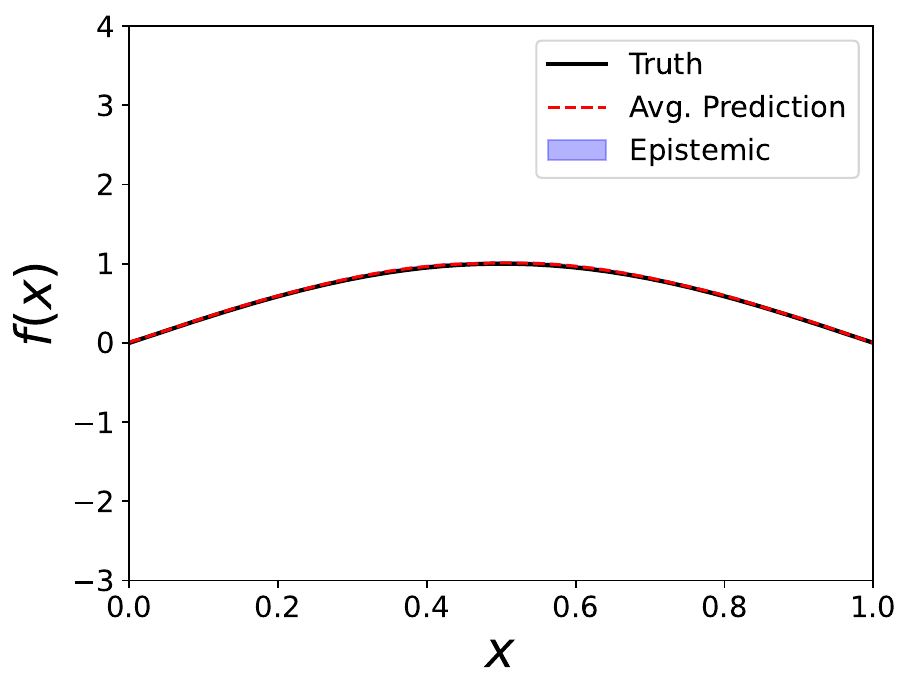} %
   \includegraphics[width=0.3\textwidth]{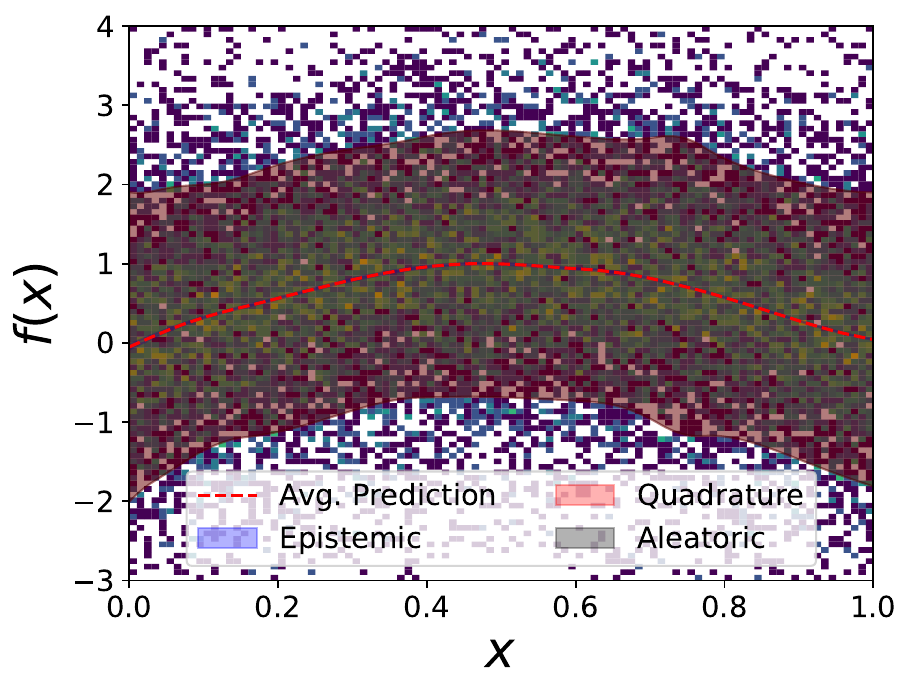} %
    \includegraphics[width=0.3\textwidth]{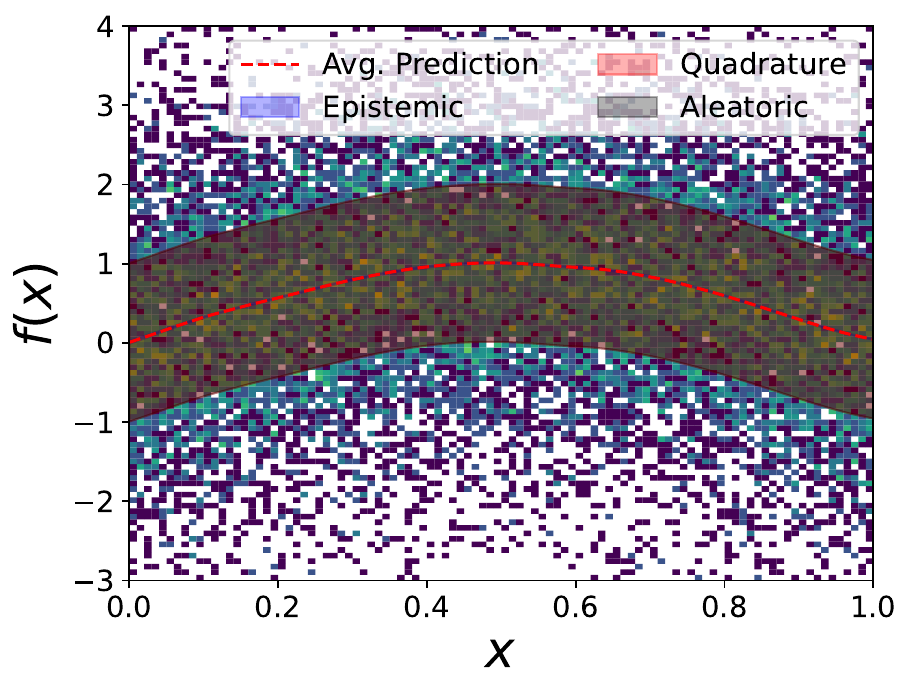} \\
    \includegraphics[width=0.3\textwidth]{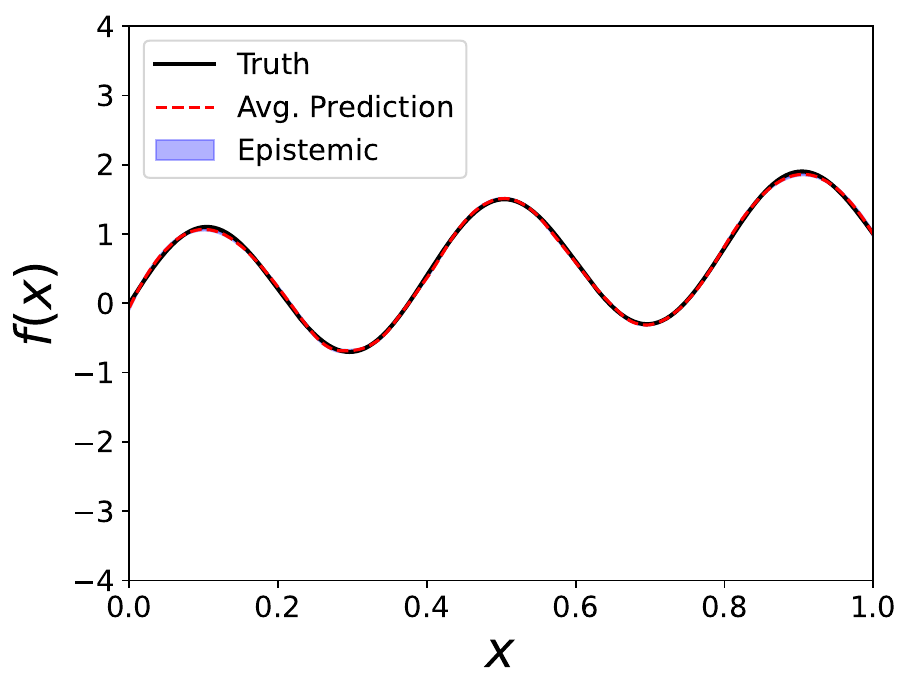} %
   \includegraphics[width=0.3\textwidth]{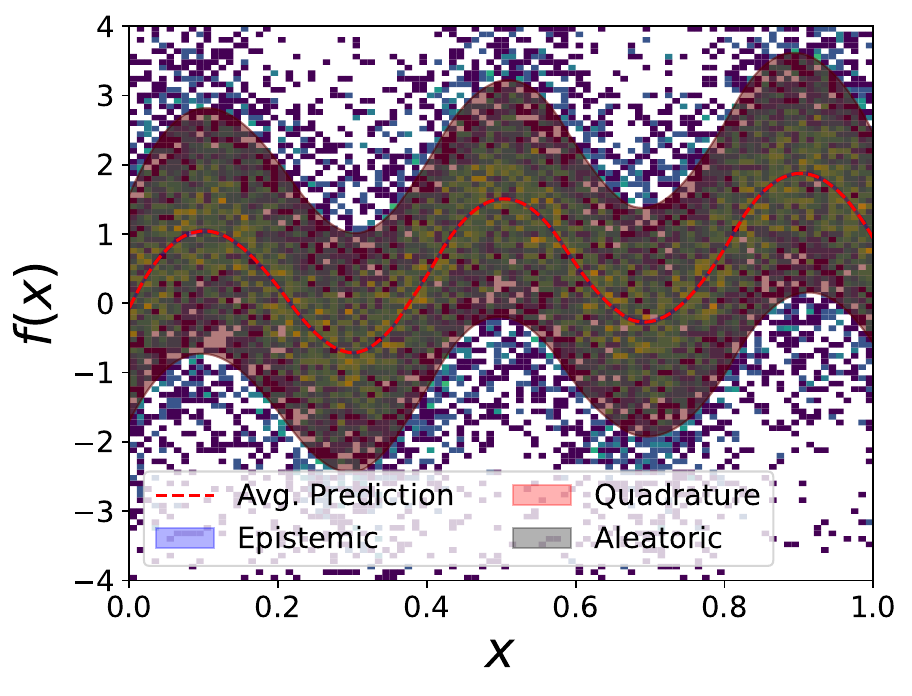} %
   \includegraphics[width=0.3\textwidth]{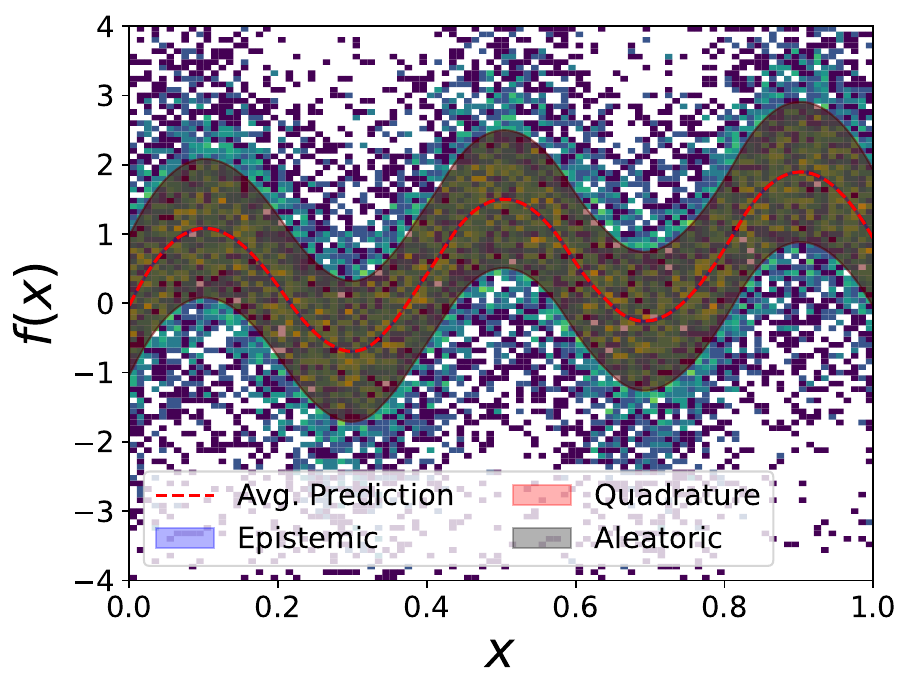} \\
    \includegraphics[width=0.3\textwidth]{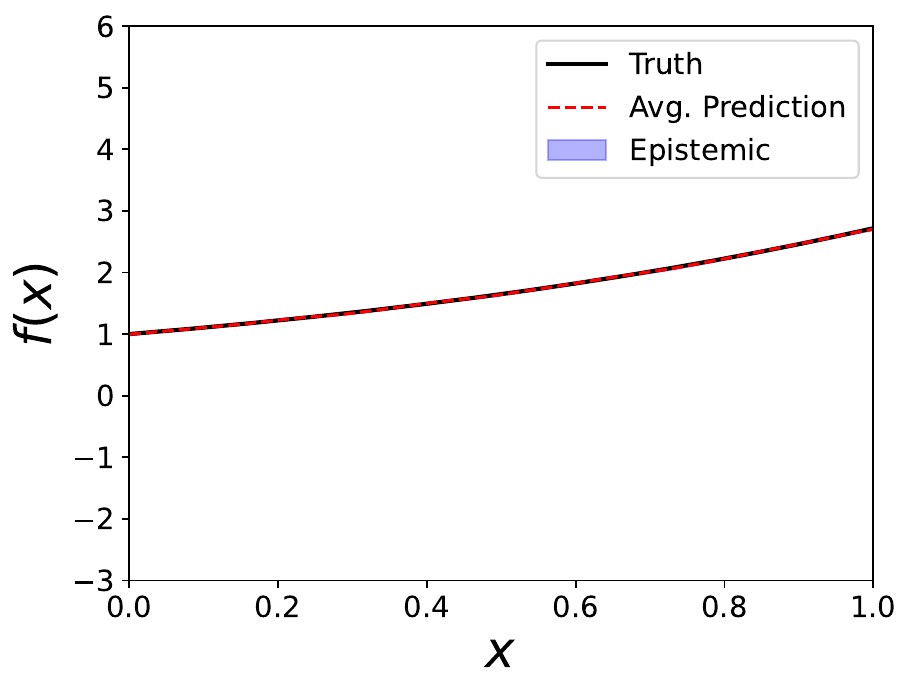} %
   \includegraphics[width=0.3\textwidth]{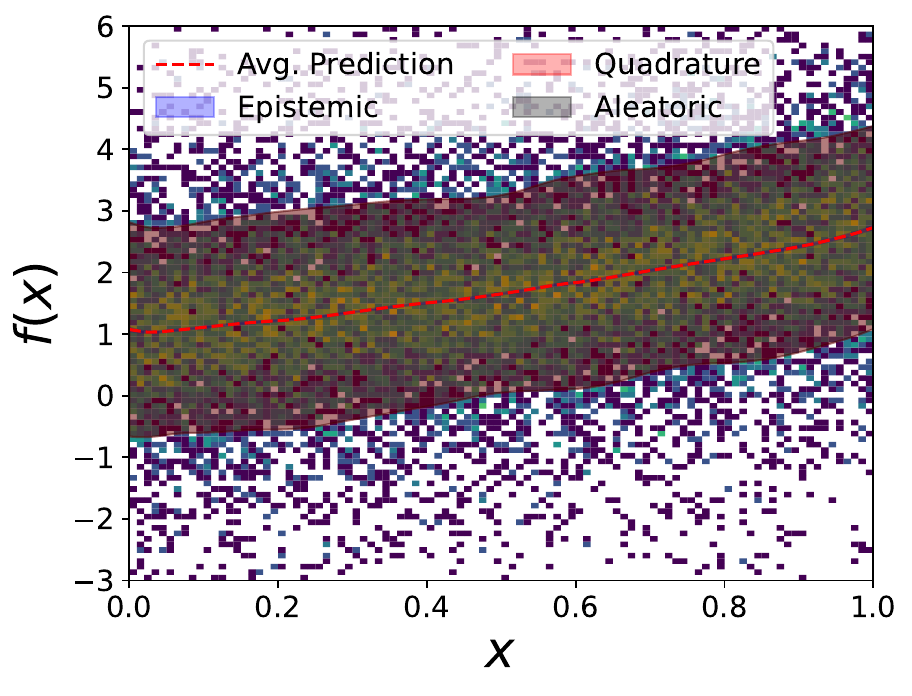} %
   \includegraphics[width=0.3\textwidth]{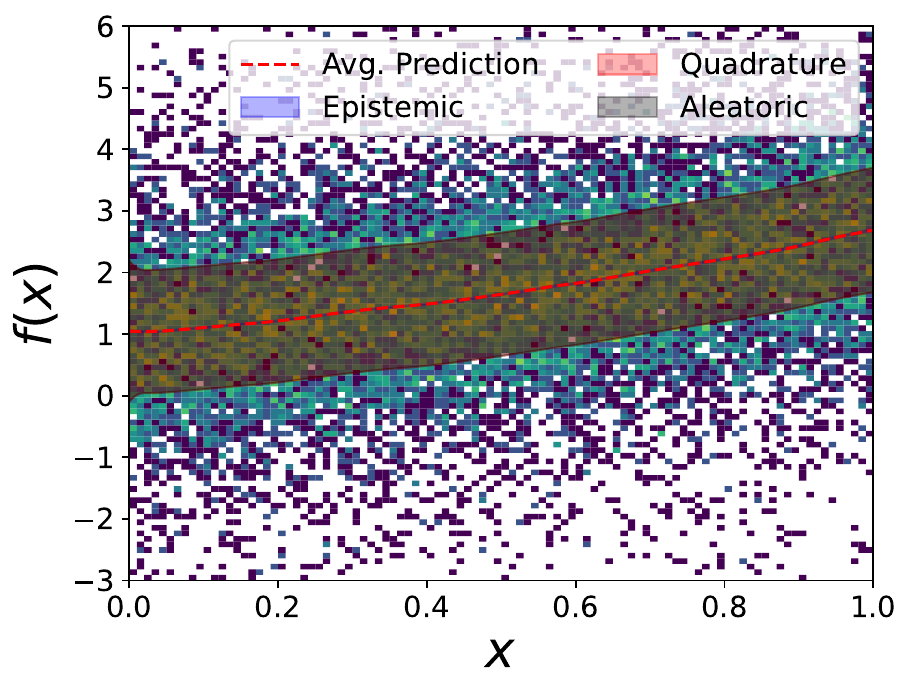} \\
    \caption{\textbf{One Dimensional Fits:} Underlying function fitted with Bayesian ReLUKAN (left column), noised function fit with a Gaussian likelihood (center column) and noise function fit with Student-t likelihood (right column). Noise is sampled from a Student-t distribution with $\nu=3$. Note in all cases, the epistemic uncertainty is very small and not visible. This is in agreement with the true inaccuracy, measured as the difference between truth and mean prediction, being almost overlapping. The aleatoric term is captured with high fidelity under both likelihood assumptions.}
    \label{fig:1D_fits}
\end{figure}

We also define the ``noised'' version of these functions, in which we introduce a stochastic term drawing samples from a Student-t distribution with $\nu = 3$. We then perform a series of fits under different likelihoods, namely a Gaussian likelihood, the center column of Fig. \ref{fig:1D_fits} and a Student-t likelihood, the rightmost column of Fig. \ref{fig:1D_fits}. The true underlying function is shown in the leftmost column. The network structure used is summarized in Tab. \ref{tab:specs} for all three functions.

From inspection of Fig. \ref{fig:1D_fits}, it can be seen that under the presence of strong outliers, \textit{i.e.}, long tails, the Student-t likelihood less affected, correctly recovering the scale parameter ($\sigma \sim 1.0$) of the core distribution, along with the correct $\nu$ value. The Gaussian likelihood under such outliers tends to overestimate the scale parameter. The results are summarized in Tab. \ref{tab:1D_metrics}. %
 Note that, in general, epistemic uncertainty decreases as the number of samples increases. In our studies, we use relatively high-statistics datasets, which allows the model to achieve low values of epistemic uncertainty across the domain.

\begin{table}[h]
    \centering
    \begin{tabular}{c c c c c}
    \toprule
   Function & Likelihood & $\sigma_{avg.}$ & 95\% Quantile Range & $\hat{\nu}$ \\
    \midrule
     $f_1$   & Gaussian & 1.717  & 1.627 - 1.871 & N/A \\
     $f_2$   & Gaussian & 1.717 & 1.575 - 1.816 & N/A \\
     $f_3$   & Gaussian & 1.698 & 1.603  - 1.776 & N/A \\
     $f_1$   & Student-t & 0.999 & 0.994 - 1.000 & 2.97 \\
     $f_2$   & Student-t & 1.005 & 0.994 - 1.1018 & 3.05 \\
     $f_3$   & Student-t & 1.003 & 1.000 - 1.008 & 3.02 \\
     \bottomrule
    \end{tabular}
    \caption{\textbf{Validation of Aleatoric Uncertainty on One-Dimensional Fits:} One-dimensional function fits with noise contribution from a Student-t distribution with $\nu = 3$. The functions are fit with two different likelihoods, namely a Gaussian and Student-t. Notice the Student-t likelihood is more robust to outliers by design, correctly recovering the scale and $\nu$ parameters ($\sigma \sim 1$ and $\hat{\nu} \sim 3$). The Gaussian likelihood tends to over estimate the scale parameter to reflect the empirical $\sigma$ value.}
    \label{tab:1D_metrics}
\end{table}


\subsection{Partial Differential Equations}

Given the model's ability to extract both epistemic and aleatoric uncertainties on the one-dimensional fits shown prior, we extend this space of solutions to Stochastic Partial Differential Equations (SPDEs). We define toy problems following the equations used in other works such as \cite{rigas2024,howard2024,shukla2024,liu2024kan,patra2024physicsinformedkolmogorovarnoldneural}. Specifically, working with the two-dimensional Poisson and Helmholtz equations.
We add a stochastic term ($f$ in Eq.\ref{eq:noise}) to validate the ability of the surrogate model to learn the aleatoric component, specifically under the case of a functional form. At each iteration of training, the noise over the solution space is sampled randomly. 

\begin{equation}\label{eq:noise}
    f = N\left(0,(|x|\sigma)^2\right), \; \sigma = 0.1
\end{equation}

This is more aligned with real-world scenarios, in which measurement devices may inherently contain a functional dependence on the amount of noise measured. In both cases, we fit a Gaussian likelihood to our function, leaving the inclusion of Student-t likelihoods for future works involving real-world applications.

\paragraph{Stochastic Poisson Equation}

A two-dimensional stochastic Poisson equation is given by:

\begin{equation}\label{eq:poisson_eq}
    \begin{aligned}
             \nabla^2 u + & 2\pi^2\sin(\pi x)\sin(\pi y) + f = 0 \\
             & \\
             & (x,y)  \in [-1,1] \times [-1,1] \\ 
             & \\
             & \begin{cases}
                u(-1,y) = 0 \\
                u(1,y) = 0 \\
                u(x,-1) = 0 \\
                u(x,1) = 0 \\
            \end{cases} \\
            & \\
            & u(x,y) = \sin(\pi x) \sin(\pi y)
    \end{aligned}
\end{equation}

We define a $64\times64$ grid over the solution space for training and use the same hyperparameter setup across all three models, \textit{i.e.}, the ReLUKAN, HR-ReLUKAN and the Bayesian HR-ReLUKAN. Specifically, we set the width of each network as [2,2,1], with a grid of 5 and a $k$ value of 3 (summarized in Tab. \ref{tab:specs}). The order used for both higher order networks is 4. All models are trained for $60k$ iterations using the Adam optimizer with a learning rate of $10^{-3}$. For training the deterministic KANs, the loss function is given by Eq. \ref{eq:mse_loss}, in which $\alpha=5\times 10^{-2}$. $\mathcal{L}_{pde.}$ corresponds to the loss contribution over the interior of the function, and $\mathcal{L}_{bc.}$ corresponds to the boundary conditions.

\begin{equation}\label{eq:mse_loss}
    \mathcal{L} = \alpha \mathcal{L}_{pde.} + \mathcal{L}_{bc.}
\end{equation}

Where mean squared error (MSE) is used between the second order derivatives on $\hat{u}$ from the network and the driving function, and on the boundary condition values defined prior. The loss function for the Bayesian networks is given by Eq. \ref{eq:bayes_loss}, in which we fit a Gaussian likelihood over both the boundary interior and boundary conditions of the PDE in combination, and scale the KL contribution with $\beta=10^{-3}$. At inference, we sample the posterior distributions over the basis functions $5k$ times to provide accurate estimations of both the aleatoric and epistemic uncertainties.

\begin{equation}\label{eq:bayes_loss}
    \mathcal{L}_{Bayes.} = \mathcal{L}(\bold{u} | \bold{x},\bold{\hat{u}}_{\phi_\bold{W}},\boldsymbol{\hat{\sigma}}_{\phi_\bold{W}})_{pde.} + \mathcal{L}(\bold{u} | \bold{x},\bold{\hat{u}}_{\phi_\bold{W}},\boldsymbol{\hat{\sigma}}_{\phi_\bold{W}})_{bc.} + \beta \mathcal{L}_{KL.}
\end{equation}

\begin{figure}[h]
    \centering
    \includegraphics[width=\textwidth]{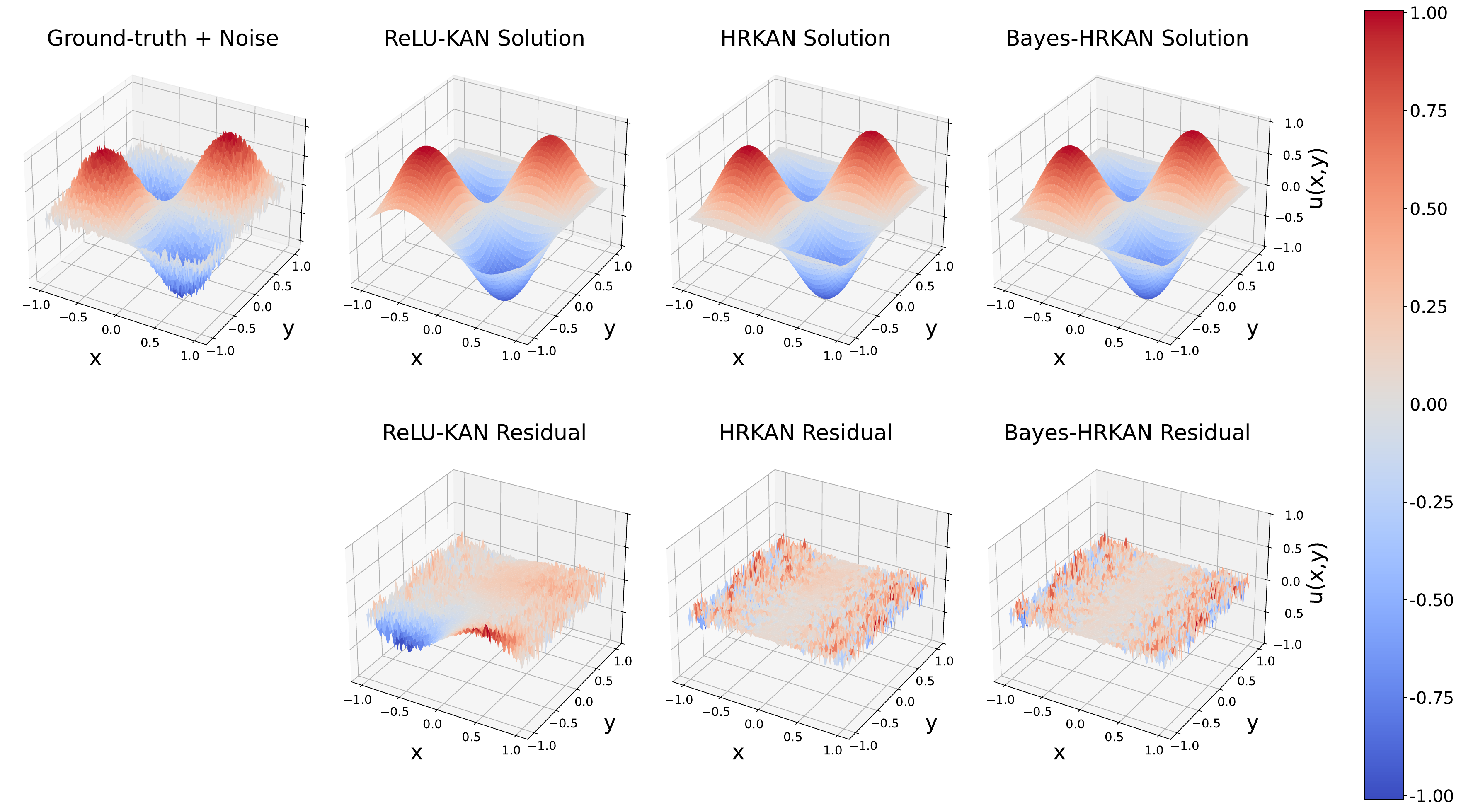} \\
    \caption{\textbf{Poisson's Equation with Gaussian Noise:} 
    Three-dimensional representations of the fitting capability of the ReLU-KAN (top row, second column), the HRKAN (top row, third column) and the Bayesian HRKAN (top row, fourth column) on the Poisson equation (top left). The residuals ($u - \hat{u}$) for each model are shown directly below and highly resemble the noise contribution, implying accurate fits towards the average values of the function. 
    }
    \label{fig:poisson_fits_3D}
\end{figure}

Fig. \ref{fig:poisson_fits_3D} depicts the ground truth with the addition of noise (top row, left), along with the resulting fits of the ReLU-KAN (top row, second column), the HRKAN (top row, third column) and the Bayesian HRKAN (top row, fourth column). The residual plots associated with each plot are shown below. Note that the residuals in all three cases highly resemble the stochastic contribution, showing that all three methods fit the average behavior well. The higher-order models are able to capture the complexity of the landscape to a higher degree. We utilize the MSE over the testing grid to evaluate the fitting performance numerically in Tab. \ref{tab:poisson_metrics}, in which it can be seen the Bayesian-HRKAN is able to obtain the same fitting results as its deterministic counterpart (HRKAN). We also report the computing time needed to train each algorithm, in which the Bayesian algorithm requires significantly more time due to the complexity of the KL computation for each individual layer.

\begin{table}[!]
    \centering
    \begin{tabular}{c c c c}
    \toprule
   Method & MSE & STD & Training Time (s) \\
    \midrule
     ReLU-KAN    & 0.007 & 0.014 & 782 \\
     HR-ReLU-KAN    & 0.004 & 0.007 & 925 \\
     Bayesian HR-ReLU-KAN    & 0.004 & 0.007 & 6290 \\
     \bottomrule
    \end{tabular}
    \caption{\textbf{Poisson Performance Averages:} Summary of performance metrics over the solution space of the 2D Poisson equation with the addition of noise.}
    \label{tab:poisson_metrics}
\end{table}

\begin{figure}[b]
    \centering
    \includegraphics[width=\textwidth]{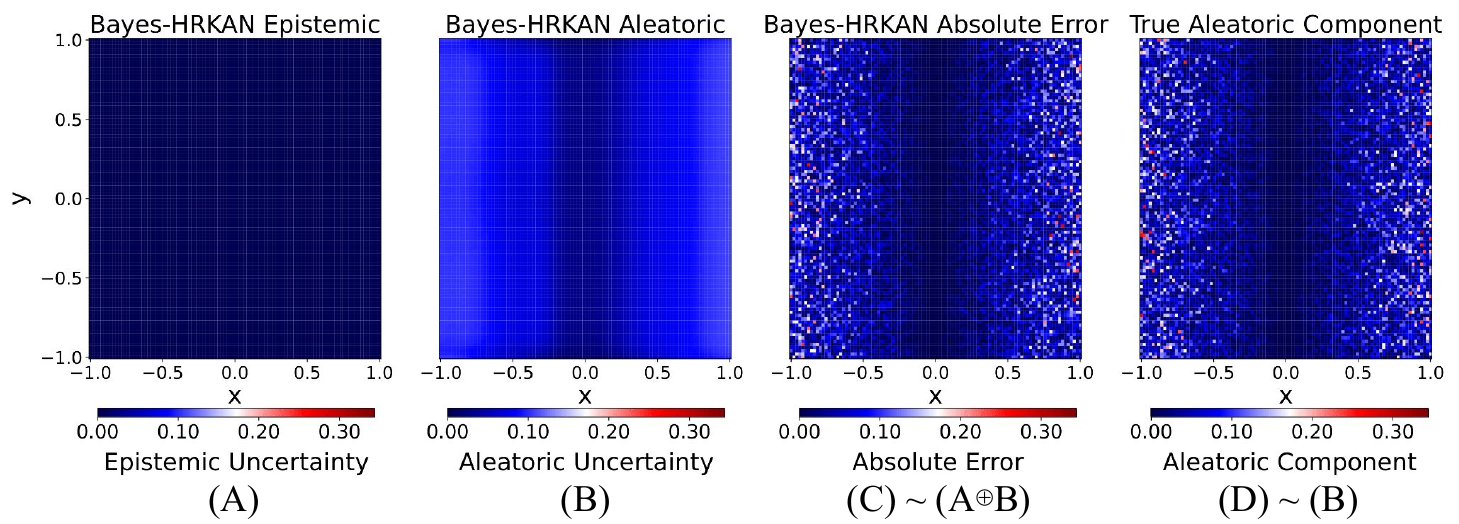}
    \caption{\textbf{Validation of Uncertainty on Poisson's Equation}: Epistemic uncertainty (A), aleatoric uncertainty (B), absolute error (C), and the true aleatoric component (absolute value) are shown (D). The epistemic uncertainty quantifies the deviation of the learned function from the true mean, with an average value of $ \bar{\sigma}_{epi.} = 0.0035 \pm 0.0001$. The accuracy of the learned aleatoric component can be observed by comparing the second and fourth plots. Note that the two rightmost plots depict individual samples rather than average behavior.}
    \label{fig:poiss_uncertainty}
\end{figure}

We also wish to validate the uncertainty produced by our network. Fig. \ref{fig:poiss_uncertainty} shows the epistemic uncertainty (left most, A), aleatoric uncertainty (second from left, B), absolute error (third from left, C) and the absolute value of the true aleatoric component (right most, D) following Eq. \ref{eq:noise}. Note that the epistemic uncertainty is a direct measurement of the learned functions deviation from the true mean.
 One should also notice the accuracy in the learned aleatoric component through comparison of the second (B) and fourth plots (D). Specifically, the learned functional dependency, that is utilized as a closure test. Note that the two right-most plots are individual samples and not average behaviors of the stochastic contribution.

\paragraph{Stochastic Helmholtz Equation}

A two-dimensional stochastic Helmholtz equation is given by:

\begin{equation}\label{eq:helmholtz_eq}
    \begin{aligned}
             \nabla^2 u + k^2 u  -[k -(a_1 & \pi)^2 - (a_2 \pi)^2] \sin(a_1\pi x)\sin(a_2\pi y) + f = 0 \\
             & \\
             & (x,y)  \in [-1,1] \times [-1,1] \\ 
             & \\
             & \begin{cases}
                u(-1,y) = 0 \\
                u(1,y) = 0 \\
                u(x,-1) = 0 \\
                u(x,1) = 0 \\
            \end{cases} \\
            & \\
             u(x,y) = \sin( & a_1  \pi x)\sin(a_2\pi y), \; a_1 = 1, a_2 = 2, k=1
    \end{aligned}
\end{equation}

\begin{figure}[h]
    \centering
    \includegraphics[width=\textwidth]{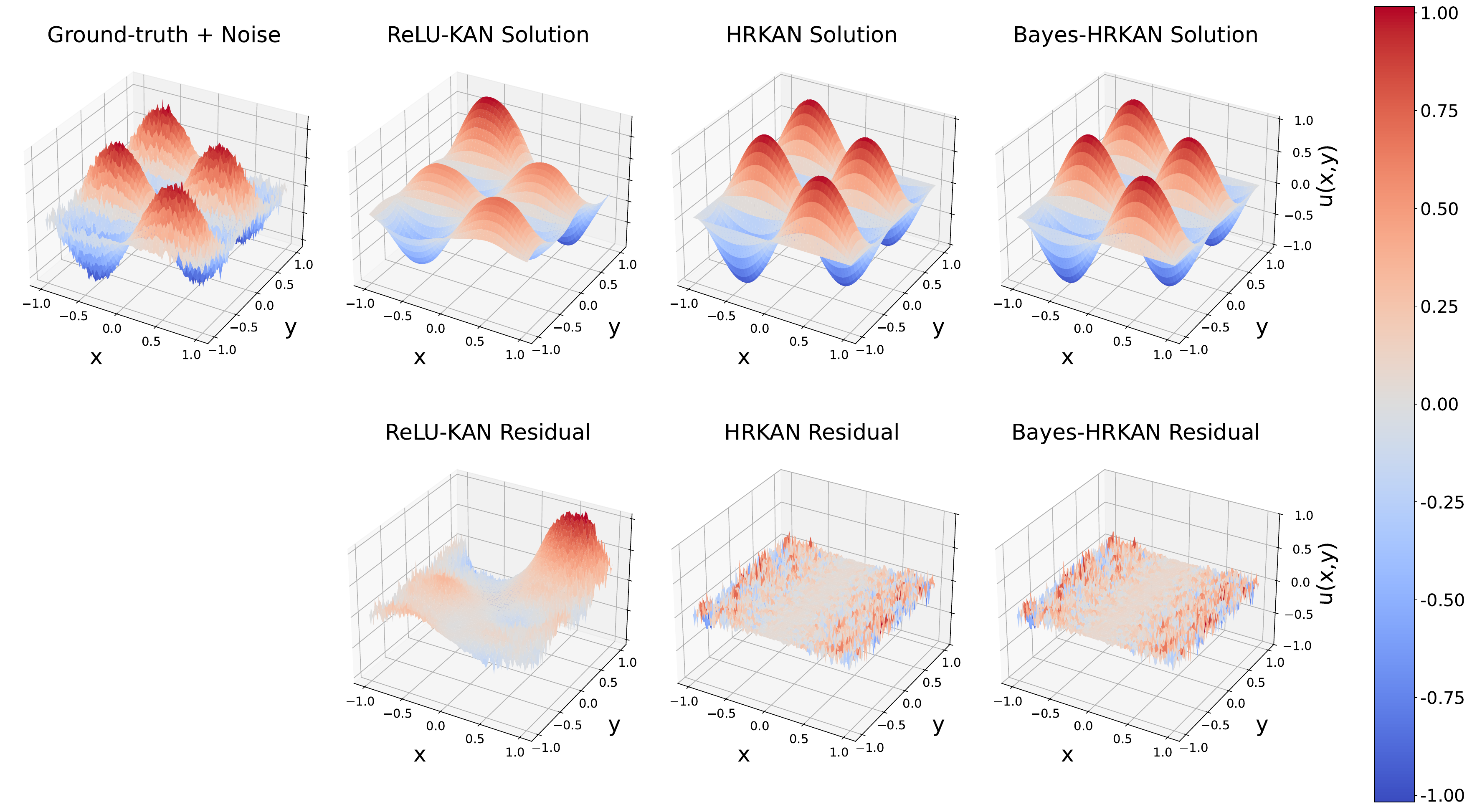} \\
    \caption{\textbf{Helmholtz's Equation with Gaussian Noise:} Three-dimensional representations of the fitting capability of the ReLU-KAN (top row, second column), the HRKAN (top row, third column) and the Bayesian HRKAN (top row, fourth column) on the Helmholtz equation (top left). The residuals ($u - \hat{u}$) for each model are shown directly below and highly resemble the noise contribution, implying accurate fits towards the average values of the function. The learned solution space is with $a_1 = 1, a_2 = 2, k=1$.}
    \label{fig:helmholtz_v1}
\end{figure}

We define a $256\times256$ grid over the solution space for training and use the same hyperparameter setup across all three models, \textit{i.e.}, the ReLUKAN, HR-ReLUKAN and the Bayesian HR-ReLUKAN. \footnote{Larger grid sizes allow better convergence on more complex functions.} Specifically, we set the width of each network as [2,2,1], with a grid of 10 and a $k$ value of 3 (summarized in Tab. \ref{tab:specs}). The order used for both higher-order networks is 4. All models are trained for $60k$ iterations. For training the deterministic KANs, the loss function is given by Eq. \ref{eq:mse_loss}, in which $\alpha=5\times 10^{-2}$, and the loss function for the Bayesian method is given by Eq. \ref{eq:bayes_loss}. Note that the loss function in this case will encompass the second order derivatives of $\hat{u}$, the function $\hat{u}$ itself, and the driving function. Note that unlike the Poisson equation shown prior, the magnitude of the driving function is relatively large (in relation to the magnitude of the function $u(x,y)$), making the learning of the aleatoric component through the Gaussian likelihood challenging. In order to reduce the deviation from the true mean, the surrogate model tends to push uncertainties towards unary values to allow the functional KAN to converge. As a result, the parameters of the Adam optimizer saturate and must be ``reset'' in order to provide better convergence. In \cite{asadi2023resettingoptimizerdeeprl}, the authors show that in the context of Reinforcement Learning (RL), the debiasing quantities of Adam quickly saturate, implying the debiasing steps will have no effect on the overall update and ultimately plague the convergence. We observe similar behavior and therefore deploy a training scheme of resetting the optimizer state every $10k$ iterations of training, in which the final $10k$ the learning rate is also dropped by an order of magnitude to $10^{-4}$. We find this to be crucial for the aleatoric component to  correctly converge under the Gaussian likelihood and can be numerically validated through the loss function, \textit{i.e.}, the likelihood becomes negative. Fig. \ref{fig:helmholtz_v1} shows the learned solution surface for the three models, along with their residuals for an individual sample.

We summarize the performance numerically in Tab. \ref{tab:helmholtz_metrics}, in which we note the improved performance of the higher-order basis functions on more complex solution spaces. Confirming the results observed in \cite{SO_2024}. It should also be noted the $\sim$ identical performance of the Bayesian and deterministic models by design. \footnote{Differences exist with the inclusion of more decimals, however we round to the first significant figure of the uncertainty profile over the solution space.}

\begin{table}[!]
    \centering
    \begin{tabular}{c c c c}
    \toprule
   Method & MSE & STD & Time (s) \\
    \midrule
     ReLU-KAN    & 0.052 & 0.129 & 2184 \\
     HR-ReLU-KAN    & 0.003 & 0.007 & 2637 \\
     Bayesian HR-ReLU-KAN    & 0.003 & 0.007 & 7714 \\
     \bottomrule
    \end{tabular}
    \caption{\textbf{Helmholtz with Noise Performance Averages:} Summary of performance metrics over the solution space of the 2D Helmholtz equation with the addition of noise.}
    \label{tab:helmholtz_metrics}
\end{table}

\begin{figure}
    \centering
    \includegraphics[width=\textwidth]{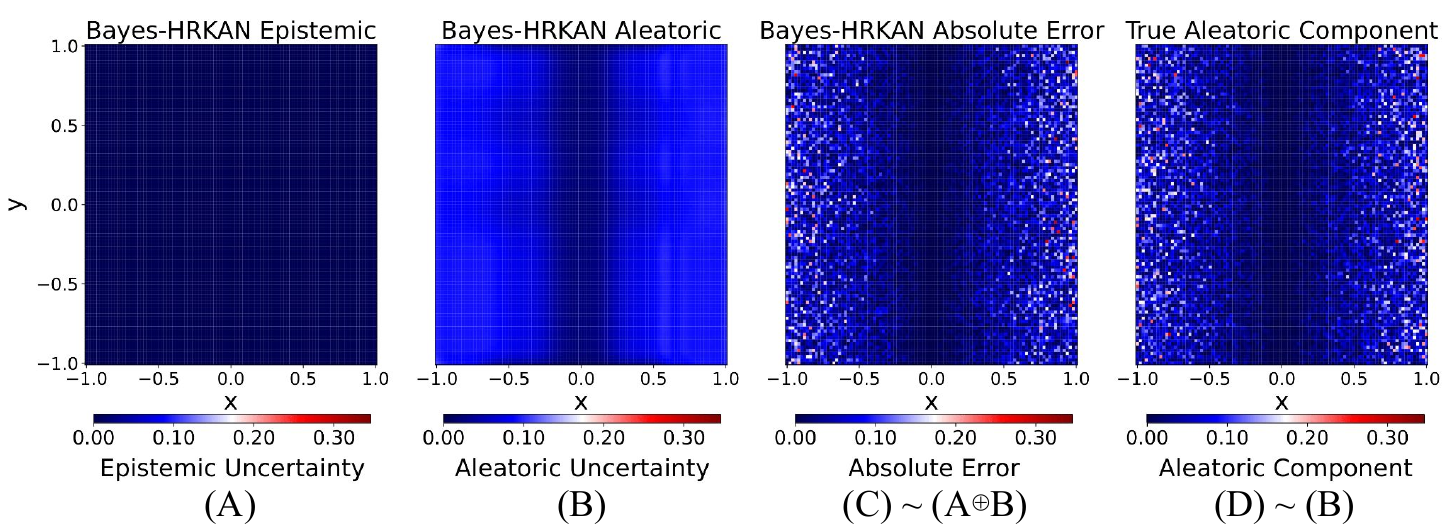} %
    \caption{\textbf{Validation of Uncertainty on Helmholtz's Equation}: Epistemic uncertainty (A), aleatoric uncertainty (B), absolute error (C), and the true aleatoric component (absolute value) are shown (D). The epistemic uncertainty quantifies the deviation of the learned function from the true mean, with an average value of $\bar{\sigma}_{epi.} = 0.0033 \pm 0.0001$. The accuracy of the learned aleatoric component can be observed by comparing the second and fourth plots. Note that the two rightmost plots depict individual samples rather than average behavior.}
    \label{fig:helmholtz_uncertainty}
\end{figure}

As done prior, we also wish to validate the uncertainty learned uncertainty profiles. Fig. \ref{fig:helmholtz_uncertainty} shows the epistemic uncertainty (left most, A), aleatoric uncertainty (second from left, B), absolute error (third from left, C) and the absolute value of the true aleatoric component (right most, D) following Eq. \ref{eq:noise}. Note that the epistemic uncertainty is a direct measurement of the learned functions deviation from the true mean, providing an average value of $\bar{\sigma}_{epi.} = 0.0033 \pm 0.0001$. One should also notice the accuracy in the learned aleatoric component through comparison of the second (B) and fourth plots (D). Note that the two right-most plots are individual samples and not average behaviors of the stochastic contribution.

\paragraph{Training and Inference Specifications}

All training and inference is done using a single NVIDIA-A40 GPU with PyTorch 2.4.0 and CUDA 12.1. Tab. \ref{tab:specs} describes the hyperparameter setup during training and inference, along with the computational overhead required by sampling.

\begin{table}[!]
    \centering
    \scalebox{0.85}{
    \begin{tabular}{c c c c c c c c}
    \toprule
        Equation &  Width & Grid  & k & Order & Test Grid (PDE) & Inference Samples & VRAM Usage (Inference) \\
        \midrule
        $f_1$ &  [1,1] & 5 & 3 & 2 & N/A & $10k$ & $\sim 24 GB$\\
        $f_2$ &  [1,1] & 5 & 3 & 2 & N/A & $10k$ & $\sim 24 GB$ \\
        $f_3$ &  [1,1] & 5 & 3 & 2 & N/A & $10k$ & $\sim 24 GB$ \\
        Poisson &  [2,2,1] & 5 & 3 & 4 & $100 \times 100$ & $5k$ & $\sim 16 GB$ \\
        Helmholtz & [2,2,1] & 10 & 3 & 4 & $100 \times 100$ & $5k$ & $\sim 26 GB$ \\
        \bottomrule
    \end{tabular}}
    \caption{\textbf{Training Specifications:} All trainings are done using a single NVIDIA-A40 GPU with 48GB of VRAM, PyTorch 2.4.0 and CUDA 12.1.}
    \label{tab:specs}
\end{table}

\paragraph{Limitations} 

As discussed prior, the fitting of the aleatoric term introduces additional complexity into the optimization scheme. Specifically, on PDEs as the magnitude of the driving function grows being that we cannot directly scale our inputs. There exists only empirical remedies as solution complexity grows, in which we find resetting the optimizer state sufficient to perturb the aleatoric surrogate model out of local minima. Moreover, the selection of the hyperparameters of the Bayesian-HRKAN itself can be finicky and we find KANs in general are generally more tricky to train than traditional neural networks. It should also be noted that the propagation of uncertainty through higher order derivatives can be difficult at the implementation level to retain connected computation graphs. This is also not extremely generalizable and must be configured equation by equation depending on the functional form, \textit{i.e.}, combinations of $n^{th}$ order derivatives.

\section{Current Trends, Future Directions, and Broader Impacts
}\label{sec:Impacts}

While KANs hold significant promise, they are a relatively new development in deep learning, and ongoing research aims to further explore their capabilities and limitations.

In \cite{liu2024kan_2}, the authors present KAN 2.0, a framework that integrates KANs with scientific discovery, emphasizing three core aspects: identifying relevant features, revealing modular structures, and deriving symbolic formulas. This bidirectional approach both integrates scientific knowledge into KANs and extracts insights from them, introducing tools like MultKAN (with multiplication nodes), a KAN compiler, and a tree converter, demonstrating its potential to uncover physical laws such as conserved quantities and symmetries, pushing the boundaries of AI and scientific exploration.
In \cite{koenig2024kan}, the authors propose KAN-ODEs, where KANs are introduced as an alternative to MLPs for data-driven modeling, forming the backbone of a Neural Ordinary Differential Equation framework. KANs are employed to efficiently model time-dependent and grid-sensitive dynamical systems, offering advantages such as faster scaling, fewer parameters, and improved interpretability compared to traditional MLP-based Neural ODEs, making them particularly suited for scientific machine learning tasks.
In \cite{patra2024physicsinformedkolmogorovarnoldneural}, the authors introduce Physics-Informed Kolmogorov-Arnold Networks (PIKAN), which leverage the Kolmogorov-Arnold theorem to solve differential equations more efficiently than traditional deep neural networks. PIKAN minimizes the need for deep layers and reduces computational costs while maintaining high accuracy.
In \cite{abueidda2024deepokan}, the authors introduce DeepOKAN, a novel neural operator that leverages KANs and Gaussian radial basis functions for improved approximation in complex engineering scenario. DeepOKAN achieves superior predictive accuracy in mechanics problems such as 1D sinusoidal waves, 2D orthotropic elasticity, and transient Poisson’s problems, outperforming conventional DeepONets \cite{lu2021learning_deeponet} in both training loss and prediction accuracy, offering a promising improvement for neural operators in engineering design.
In \cite{xu2024kolmogorov}, the authors explore KANs for time-series forecasting, focusing on detecting concept drift and improving predictive performance and interpretability. This approach facilitates adaptive forecasting models in predictive analytics, enabling more responsive and accurate predictions in dynamic environments.

At present, KANs appear to be more suitable for tasks involving relatively low-dimensional data, excelling in problems with continuous functions but facing challenges with high-dimensional, nonlinear datasets and large-scale applications.
Despite these limitations, ongoing research explores various applications, such as KANs in continuous optimal control problems (\textit{e.g.}, \cite{aghaei2024kantrol}, applying KANtrol to the 2D heat equation) and core temperature estimation for lithium-ion batteries without surface sensor feedback \cite{karnehm2024core}.
The proposed Bayesian KAN can be highly valuable for uncertainty quantification across multiple scenarios. A promising direction is data-driven fitting of stochastic differential equations with integrated uncertainty quantification, providing robust and interpretable models, particularly for systems where uncertainty plays a critical role in prediction and decision-making.

\section{Conclusion}

We have introduced the first method of uncertainty quantification in the domain of KANs, specifically focusing on (Higher Order) ReLUKANs \cite{qiu2024relu,SO_2024} to enhance computational efficiency given the overhead of Bayesian methods. Our method is general and can be translated to various other basis functions given a method of defining posteriors over their parameter space. Our method has been validated on simple one-dimensional functions, demonstrating accurate fitting procedures and precise representations of both epistemic uncertainty (true deviation from the mean) and aleatoric uncertainty (stochastic noise). This approach has been extended to the domain of SPDEs, where we construct toy scenarios for the Poisson and Helmholtz equations with the inclusion of a stochastic term. Additionally, we have demonstrated the method's ability to correctly identify functional dependencies introduced by this noise. We also identified potential challenges in optimizing the aleatoric model under large driving functions, specifically referring to large magnitudes in relation to the underlying solution and noise. In these cases, the optimizer generally pushes the aleatoric component away from
its true values. To cope with those cases, in this work we proposed solutions based on empirical approaches found in other fields \cite{asadi2023resettingoptimizerdeeprl}.

\vspace{-0.25cm}


\section*{Acknowledgments}

We thank  William  \&  Mary for supporting the work of CF and JG through CF’s start-up funding. The authors acknowledge William \& Mary Research Computing for providing computational resources and technical support that have contributed to the results reported within this article. We would also like to thank Dr. Karthik Suresh for insightful discussions surrounding this work.

\bibliographystyle{iopart-num}
\bibliography{references}

\newpage
\appendix  

\section{Epistemic Uncertainty Plots}

\begin{figure}[h]
    \centering
    \includegraphics[width=0.48\textwidth]{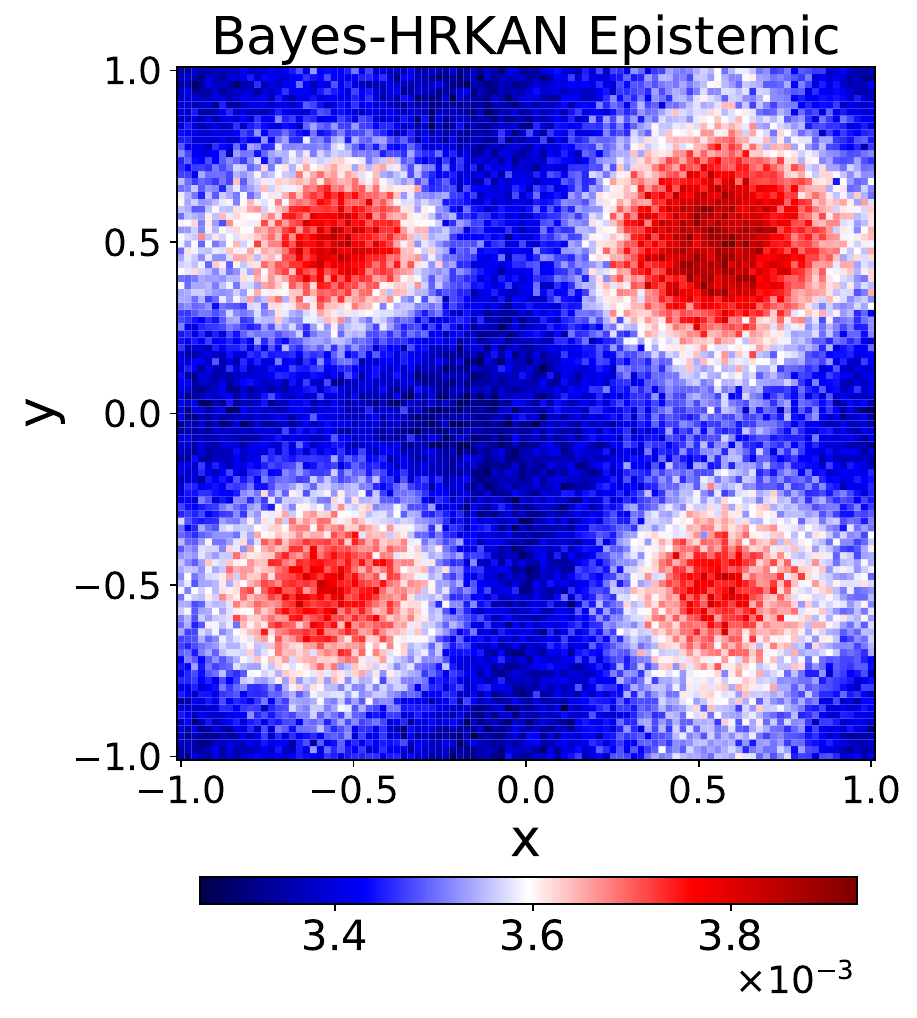} %
    \includegraphics[width=0.48\textwidth]{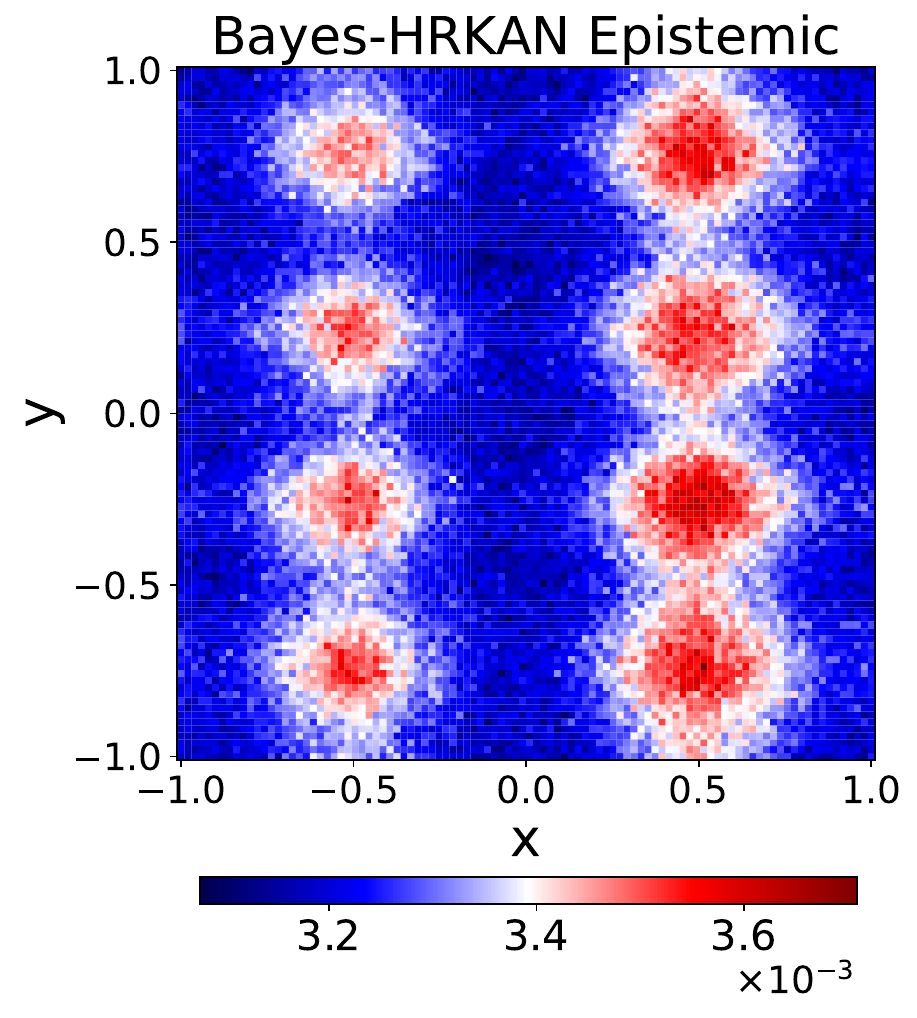}
    \caption{\textbf{Epistemic Uncertainty:} Epistemic uncertainty for models trained on the 2D Poisson equation (left), and 2D Helmholtz equation (right). The epistemic uncertainty has not been normalized to the maximum error in the learned solution (referring to Figures. \ref{fig:poiss_uncertainty}, \ref{fig:helmholtz_uncertainty}).}
    \label{fig:epi_normed}
\end{figure}

Fig. \ref{fig:epi_normed} depicts the epistemic uncertainty of the models for the 2D Poisson equation (left), and the 2D Helmholtz equation (right). Note that in comparison to Figures \ref{fig:poiss_uncertainty} and \ref{fig:helmholtz_uncertainty}, the epistemic uncertainty has not been normalized to the maximum error over the function. This allows visualization of more intricate patterns of uncertainty across the solution space, in which it can be seen the epistemic uncertainty is correlated to the magnitude of the solution. In general, the epistemic uncertainty is negligible in comparison to the error, given the high quality fit to the underlying function, \textit{i.e.}, the solution without any noise.

\clearpage

\end{document}